\documentclass[nohyperref]{article}

\usepackage{microtype}
\usepackage{graphicx}
\usepackage{subfigure}
\usepackage{booktabs} 

\usepackage{hyperref}

\usepackage[accepted]{icml2022}

\usepackage{amsmath}
\usepackage{amssymb}
\usepackage{mathtools}
\usepackage{amsthm}
\usepackage{multirow}
\usepackage{makecell}
\usepackage{xcolor}
\usepackage[capitalize,noabbrev]{cleveref}

\theoremstyle{plain}

\theoremstyle{definition}

\theoremstyle{remark}

\usepackage{dsfont}

\usepackage{pifont}

\definecolor{darkpastelgreen}{rgb}{0.13, 0.9, 0.34}
\newcommand{\cmark}{\ding{51}}%
\newcommand{\xmark}{\ding{55}}%

\usepackage[textsize=tiny]{todonotes}

\icmltitlerunning{Rethinking Attention-Model Explainability through Faithfulness Violation Test}

\newcommand{\bfstart}[1]{\noindent\textbf{#1.}}

\newcommand{\ie}{\textit{i}.\textit{e}.}
\newcommand{\eg}{\textit{e}.\textit{g}.}

\begin{document}
	
	\twocolumn[\icmltitle{Rethinking Attention-Model Explainability through Faithfulness Violation Test}



	\icmlsetsymbol{equal}{*}
	
	\begin{icmlauthorlist}
		\icmlauthor{Yibing Liu}{cityu}
		\icmlauthor{Haoliang Li}{cityu}
		\icmlauthor{Yangyang Guo}{nus}
		\icmlauthor{Chenqi Kong}{cityu}
		\icmlauthor{Jing Li}{polyu}
		\icmlauthor{Shiqi Wang}{cityu}
	\end{icmlauthorlist}
	
	\icmlaffiliation{cityu}{City University of Hong Kong, Hong Kong}
	\icmlaffiliation{polyu}{The Hong Kong Polytechnic University, Hong Kong}
	\icmlaffiliation{nus}{National University of Singapore, Singapore}
	
	\icmlcorrespondingauthor{Haoliang Li}{haoliang.li1991@gmail.com}
	
	\icmlkeywords{Attention Mechanism, Explainability, Interpretability, Faithfulness}
	
	\vskip 0.3in
	]



\printAffiliationsAndNotice{}  


\begin{abstract}

	Attention mechanisms are dominating the explainability of deep models. They produce probability distributions over the input, which are widely deemed as feature-importance indicators.
	However, in this paper, we find one critical limitation in attention explanations: weakness in identifying the polarity of feature impact.
	This would be somehow misleading -- features with higher attention weights may not faithfully contribute to model predictions; instead, they can impose suppression effects.
	With this finding, we reflect on the explainability of current attention-based techniques, such as Attention $\odot$ Gradient and LRP-based attention explanations. We first propose an actionable diagnostic methodology (henceforth \emph{faithfulness violation test}) to measure the consistency between explanation weights and the impact polarity. 
	Through the extensive experiments, we then show that most tested explanation methods are unexpectedly hindered by the faithfulness violation issue, especially the raw attention.
	Empirical analyses on the factors affecting violation issues further provide useful observations for adopting explanation methods in attention models.

\end{abstract}

\section{Introduction}
\begin{figure}
	[t]
	\centering 
	\includegraphics[width=0.89\columnwidth]{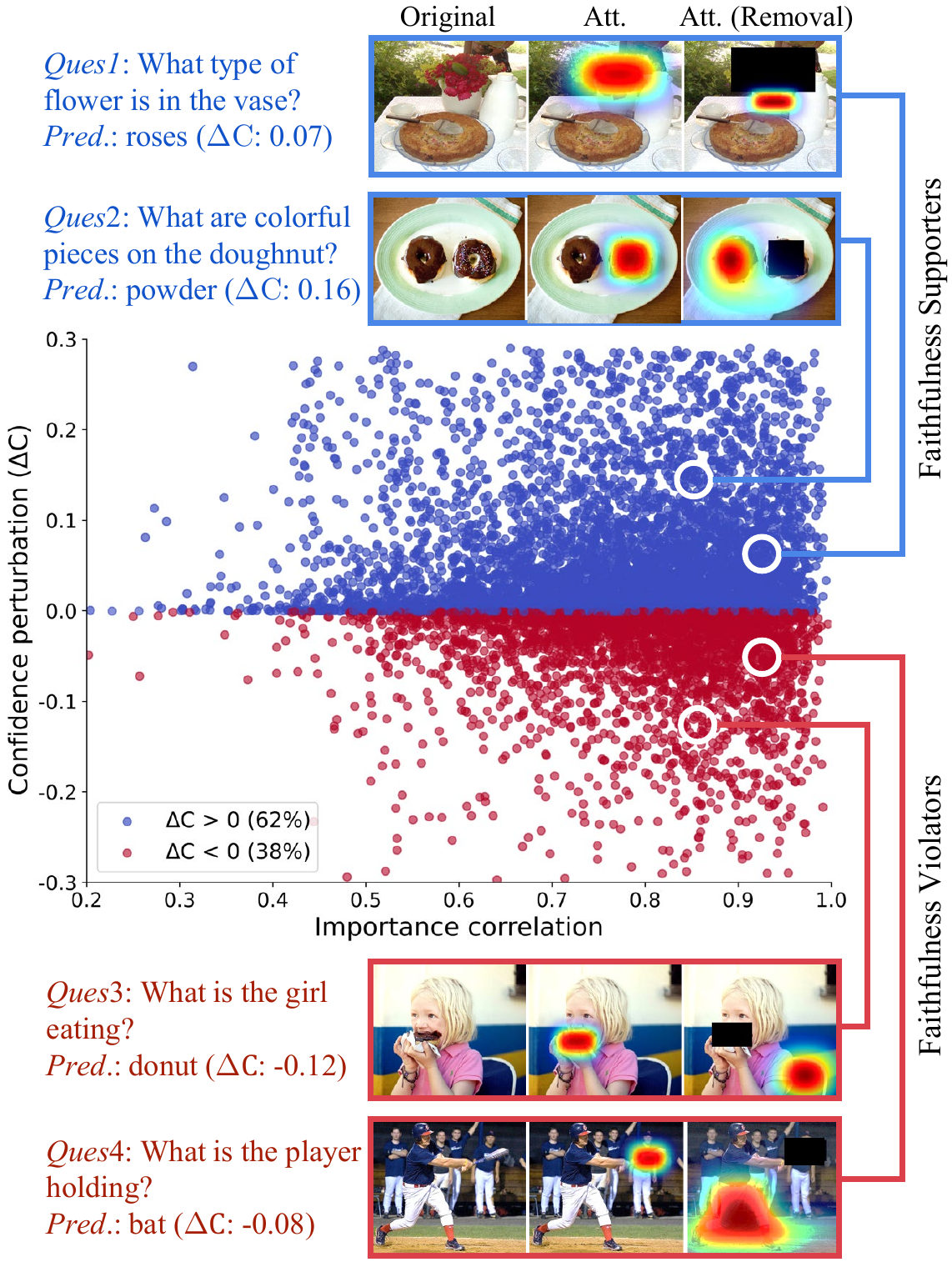} %
	\caption{Data map for attention faithfulness on BUTD model over VQA v2. The $y$-axis shows the confidence perturbation $\Delta {\rm C}$ on the original predicted label (\eg, \emph{donut}) after removing top-10\% image regions \emph{w.r.t} attention weights. $\Delta{\rm C}<0$ indicates confidence increases. The \emph{Red} points represent the ones violating \emph{polarity consistency} since image regions with the largest positive weights unexpectedly suppress model predictions, \ie, $\Delta{\rm C}<0$. The $x$-axis shows the \emph{importance correlation} for attention weights. }
	\label{intro}
\end{figure}

Recent advances in attention models~\cite{baseline:TanhAtt,baseline:Transformer(DotAtt)} have raised a pressing concern for the explainability of their black-box nature.
Prior efforts target at unraveling puzzles in inherent attention explanations, such as whether attention is directly explainable~\cite{attfinding:AttIsNotExp,attfinding:2019IsAttIntepret,attfinding:AttIsNotNotExp,transformer:Identity}, or how attention behaves~\cite{attfinding:Bert_Behavior,attfinding:AttentionsIsNotPureImportance,transformer:AttGrad_SixHeads_2019,attfinding:LSTM_Concinity}. 
In contrast, another line of work strives to devise better explanation methods for attention models~\cite{exp:IG,exp:SmoothGrad,exp:DeepLIFT,transformer:AttGrad_Chefer,transformer:IGAttGrad,attfinding:AttNorm}.
However, in this work, we find that they all lack proper evaluation in \emph{explanation  faithfulness}~\cite{faithful:DefineFaithful} -- how accurately an explanation reflects the reasons behind a model prediction. 

We argue that a faithful explanation should at least possess two properties: 1) \emph{importance correlation}: the magnitude of explanation weights should precisely reflect the importance of input components; 2) \emph{polarity consistency}: the sign of explanation weights should correctly indicate the polarity of input impact, \ie, contribution or suppression effects to the model predictions.
Existing faithfulness evaluations, however, merely focus on the first property without examining the polarity consistency~\cite{attfinding:AttIsNotExp,attfinding:AttIsNotNotExp,attfinding:2019IsAttIntepret,attfinding:DiagnosticAllProperties,faithful:ImrpoveFaithfulForNMT,faithful:ImrpoveFaithfulForTC}.
While somewhat dissatisfying, this can be heavily misleading: in cases where an explanation confuses the polarity but well correlates with importance, we may mistakenly identify inputs (\eg, image regions) with large suppression effects as the reason for model predictions.

To illustrate this point, we build a dataset map~\cite{exp:datasetmap} in Figure \ref{intro} for attention faithfulness evaluation over the Visual Question Answering (VQA) task, which distinguishes between the samples that violate {polarity consistency} (\emph{Red} points) and the ones support (\emph{Blue} points). 
{As can be seen, in numerous appealing cases} where even attention weights precisely reflect importance of image regions (\ie, high {importance correlation}), highlighted regions may not faithfully contribute to the predictions; instead, they impose suppression effects, \ie, violating {polarity consistency}.
This hence would be fallacious if we believe such highlighted image regions help the model prediction.
As such, in this paper, we argue that {polarity consistency} is essential to be a criterion for faithful explanations, and a re-assessment for {explanation faithfulness} is imperative.

To this end, we propose an actionable diagnostic methodology, \emph{violation test}, to evaluate the under-explored property, polarity consistency, in faithfulness.  
In particular, this violation test would take the largest explanation weight (in absolute value) as the representative for each explanation sample, and then examine the correctness of its corresponding polarity, \ie, whether denoting the actual direction of input impact. 
In response, if the polarity is correct, such an explanation sample would be marked as the supporter for faithfulness; otherwise, the violator.
{We should expect an explanation method passes the test with as
few violations as possible, if it faithfully reflects model decisions.}
In addition, we also compare our proposed violation test with existing faithfulness evaluation metrics (Section \ref{sec_q1:why_vio}). Through the results, we show that they all lack the ability in examining polarity consistency, which further highlights the necessity of our violation test for faithfulness evaluation.
Note that our test method is also applicable to any explanation method without modifications.




With the realization of the weakness in existing faithfulness evaluation, we instantiate our analysis on nine widely utilized explanation methods, \eg, Attention Rollout~\cite{transformer:AttFlow}.
In our broad experimental sweep over six tasks and nine datasets, we find that most analyzed explanation methods are unexpectedly hindered by the faithfulness violation issue (Section \ref{sec_q2:how_exp}). Specifically, the single-polarity methods (\eg, Raw Attention) perform the worst, while the form of gradient-based attention explanations can largely alleviate violation issue. With these findings, we further study the factors leading to the violation issue (Section \ref{section:factors}). Based on our experiments, we show that the complexity in model architectures and the capability to identify polarity can be the main factors.



To sum up, this work makes the following contributions:
\begin{itemize}
	\item We study the under-explored dimension, \emph{polarity consistency}, in the faithfulness evaluation. We propose a violation test to re-assess the faithfulness of explanation methods. By comparison with existing faithfulness evaluation metrics, we show the indispensability of our proposed violation test.
	
	\item We conduct extensive experiments with three groups of explanation methods across tasks, datasets, and model architectures. We find, consistently, most methods tested are limited by the faithfulness violation issue. In particular, raw attention explanations perform the worst, while the form of attention $\odot$ gradient can greatly alleviate the violation issue and even pass the violation test in some special cases.
	
	\item 
	We investigate the dominant factors leading to the violation issue. The results suggest that the complexity in model architectures and the capability of identifying polarity play important roles. All code for model implementation and data processing is made avaliable\footnote{\url{ https://github.com/BierOne/Attention-Faithfulness}}.
\end{itemize}


\section{Related Work}
\subsection{Model Explainability}
Explaining the internal behavior of black-box neural networks has been extensively studied in recent years. They can be grouped into several families, including but not limited to gradient-based methods~\cite{exp:DeepLIFT,exp:IG,exp:LRP,exp:SmoothGrad}, perturbation-based explanations~\cite{saliency:perturbation1,saliency:perturbation3,saliency:perturbation2}, and Shapley values~\cite{saliency:shapley1,saliency:shapley2}.



Specific to attention models, a canonical explanation tool is attention mechanisms~\cite{baseline:TanhAtt,baseline:Transformer(DotAtt)}. 
Since attention calculates a probability distribution over the input, attention weights are often compared to human vision~\cite{humanVision:HV} and regarded as a kind of importance indicator~\cite{attTask:ImageCls,attTask:NMT,attTask:VQA}.
In line with this, many works hence develop explanation methods by combining attention weights with additional attributions, such as the gradients~\cite{transformer:AttGrad_Chefer,transformer:IGAttGrad}, layer-wise relevance~\cite{transformer:LRP_Partial_2019,transformer:LRP_Chefer}, or the norm of input vectors~\cite{attfinding:AttNorm}.
On the other end of the spectrum, many studies also seek to understand the behavior of attention. For example, \citet{attfinding:LSTM_Concinity} explored why attention fails to explain LSTM models and pointed out that the similar hidden states in the encoder restrict the significance of attention weights. In contrast, another group of work investigates what patterns attention often looks at~\cite{attfinding:Bert_Behavior,attfinding:AttentionsIsNotPureImportance}. 
Unlike them, in this paper, we provide insights for attention-based explanations from the perspective of polarity consistency.

{This work is also broadly connected to the evaluation of saliency methods~\cite{saliency:SanityCheck,saliency:SanityMetrics,saliency:ShiftInvariance_ICLR2021}. Contrary to these works mostly focusing on how explanations can be arbitrarily manipulated, our polarity consistency view centers on evaluating polarity indications of explanation weights, which sheds light on another research direction in XAI.
}

\subsection{Faithfulness in Attention Models}
An explanation satisfying faithfulness should at least possess two properties: 1) importance correlation: precisely reflecting the importance of each input feature, and 2) polarity consistency: correctly identifying the impact polarity of features. Despite recent efforts on faithfulness evaluation where many metrics have been proposed~\cite{replace:benchmark,attfinding:DiagnosticAllProperties}, they merely focus on the first property without examining the polarity consistency. 

The first attempt to explore faithfulness in attention models is~\citet{attfinding:AttIsNotExp,attfinding:2019IsAttIntepret}, where they found that large attention weights do not necessarily indicate high importance on model predictions and thus drew the conclusion that attention is not faithful. An immediate response~\cite{attfinding:AttIsNotNotExp}, however, argues their experiments and presents a different claim that attention can provide faithful explanations in certain cases. Furthermore, another work compares attention weights with other explanations, such as LIME~\cite{exp:LIME,faithful:att_lime} and Shapley values~\cite{faithful:att_shapley}.
In addition to the exploration of faithfulness, some studies also propose to improve faithfulness of attention-based explanations on the specific domains~\cite{faithful:ImrpoveFaithfulForNMT,faithful:ImrpoveFaithfulForTC}. 
However, their faithfulness measurements are all performed on the importance correlation of attention-based explanations, which heavily limit the reliability of the results.

\section{Faithfulness Violation Test}
The problem we are investigating is to what extent an explanation method can correctly identify the impact polarity of input features. 
Without loss of generality, suppose we have an input $x = \{x_i\}_{i}^{N}$, where each $x_i$ represents a feature or token. A model is represented as a function $f (\cdot)$, predicting the class $\hat{y}$ for the input:  $\hat{y} = \mathop{\arg\!\max}_y f(x)_y $. Let $w$ be an explanation method, which assigns weights for each token in the input: $e_{i} = w(x_{i}), i\in [1, N]$.

The intuition behind faithfulness violation test is that features with the largest explanation weights (in absolute values) should at least demonstrate the consistent polarity, \eg, largest positive weights ought to indicate contribution effects instead of suppression. In this sense, our first step is to find the most influential feature $x^*$ (denoted by the explanation method) in the input: $x^* = \mathop{\arg\!\max}_{x_i \in x}||w(x_i)||$.
Next, to measure the actual impact of $x^*$, we perform a well-established approach -- perturbation examination~\cite{replace:Erasure_Jiwei_2016,saliency:RemOveAndRetrain_BeemKim2019,replace:benchmark} -- replacing the certain input features ($x^*$) and calculating the confidence perturbation in model predictions:
\begin{equation}
	\Delta {\rm C}(x, x^*) = f(x)_{\hat{y}} - f(x \backslash x^*)_{\hat{y}}.
\end{equation}
In particular, if the explanation weights $w(x^*)$ are highly positive, the confidence degradation ($\Delta {\rm C}(x, x^*) > 0$) should be observed after the feature replacement. Should the confidence upgrades, however, we can infer that such an explanation violates the polarity consistency. 
As such, the faithfulness violation is measured via
\begin{equation}
	{\rm Violation} = \mathds{1}_{{\rm sign}(w(x^*) \cdot \Delta {\rm C}(x, x^*)) <0}
\end{equation}

\section{Experimental Setup}


\begin{table}[t]
	\caption{Task and dataset statistics. We provide more details for each dataset in Appendix \ref{appendix:datasets}.}\smallskip
	\centering
	\resizebox{0.95\columnwidth}{!}{
		\smallskip\begin{tabular}{c c  c  c}
			\toprule
			Task & Dataset & \# Classes & \# Train \\
			\midrule
			\multirow{2}{*}{Sentiment Analysis} 
			& SST & 5 &  6,355 \\
			& Yelp & 5 &  650,000 \\
			\midrule
			\multirow{2}{*}{\makecell[c]{Topic Classification}} 
			& AGNews & 4 &  101,998 \\
			& 20News & 20 &  10,833 \\
			\midrule
			\multirow{1}{*}{Paraphrase Detection} & QQP & 2 &  327,460 \\
			\midrule
			\multirow{2}{*}{\makecell[c]{Natural Language \\Inference}} 
			& \multirow{2}{*}{SNLI} & \multirow{2}{*}{3} & \multirow{2}{*}{549,367}  \\
			&&&\\
			\midrule
			\multirow{1}{*}{Question Answering} & bAbI-1 & 20 &  10,000  \\
			\midrule
			\multirow{2}{*}{\makecell[c]{Visual Question \\Answering}} 
			& VQA 2.0 & 3130 &  411,272 \\
			& GQA & 1842 &  943,000 \\
			\bottomrule
		\end{tabular}
	}
	\label{datasets}
\end{table}

\subsection{Tasks and Datasets}
Our goal is to explore the faithfulness violation issue of attention-based explanations.
Following prior studies~\cite{attfinding:AttIsNotExp,attfinding:LSTM_Concinity}, we conduct {extensive} experiments {on} six exemplar tasks, for which attention models are widely applied. 
Besides, to ensure the reliability of our experimental results, we select nine datasets across different training data scales, class numbers, and learning targets.
We report the setup of them in Table \ref{datasets}. 

\begin{table}[t]
	\caption{The most common explanation techniques for attention models are analyzed in this paper. ${\alpha}$ denotes the raw attention weights. $||v(x)||$ represents the norm of transformed input vectors in attention module. ${\rm R}^{\alpha}$ is the Layer-wise Relevance Propagation (LRP) scores for the corresponding attention head. Appendix \ref{appendix:expmethod} provides an in-depth description for these techniques.}\smallskip
	\centering
	\resizebox{1\columnwidth}{!}{
		\smallskip\begin{tabular}{l c c}
			\toprule
			Method & Denoted & Basis \\
			\midrule
			\multicolumn{3}{l}{\emph{Generic attention-based explanation methods}} \\
			Inherent Attention Explanation & RawAtt & ${\alpha}$ \\
			
			Attention $\odot$ Gradient & AttGrad & ${\alpha} \odot \nabla\alpha$ \\
			
			Attention $\odot$ InputNorm & AttIN & ${\alpha} \odot ||v(x)||$ \\
			\midrule
			
			\multicolumn{3}{l}{\emph{Transformer-based explanation methods}} \\
			Partial LRP & PLRP & ${\rm R}^{\alpha}$ \\
			
			Attention Rollout & Rollout & ${\alpha}$ \\
			
			Transformer Attention Attribution & TransAtt & $\nabla\alpha \odot {\rm R}^{\alpha}$ \\
			
			Generic Attention Attribution & GenAtt & ${\alpha} \odot \nabla\alpha$ \\
			
			\midrule
			\multicolumn{3}{l}{\emph{Gradient-based attribution methods}} \\
			Input $\odot$ Gradient & InputGrad & ${x} \odot \nabla x$ \\
			Integrated Gradients & IG &  ${x} \odot \nabla x$ \\
			
			\bottomrule
		\end{tabular}
	}
	\label{exp_methods}
\end{table}

\subsection{Models}
In line with previous studies~\cite{attfinding:2019IsAttIntepret,attfinding:AttIsNotExp,attfinding:DiagnosticAllProperties}, we consider two commonly employed base architectures, LSTM~\cite{tech:LSTM} and CNN~\cite{tech:CNN}, for the feature extraction on the standard NLP tasks, \eg, \emph{Sentiment Analysis}. 
Since our goal is to analyze attention-based explanations, we incorporate two kinds of attention mechanisms, additive attention (\emph{TanhAtt})~\cite{baseline:TanhAtt} and scaled dot-product attention (\emph{DotAtt})~\cite{baseline:Transformer(DotAtt)}, for the above architectures as well. As such, we experiment with four general attention models here: \textbf{LSTM+TanhAtt}, \textbf{LSTM+DotAtt}, \textbf{CNN+TanhAtt}, and \textbf{CNN+DotAtt}, all of which is initialized with the GloVe~\cite{tech:Glove} embeddings.


We utilize two typical transformer-based models, \textbf{VisualBERT}~\cite{baseline:visualBERT} and \textbf{LXMERT}~\cite{baseline:vqa_lxmert}, on the VQA task since they are representatives of self-attention and cross-modal attention mechanisms, respectively.
In addition, we also use two widely adopted general attention models, \textbf{StrongBaseline}~\cite{baseline:vqa_strong} and \textbf{BUTD}~\cite{baseline:vqa_UpDn}, on the VQA task for a thorough comparison. 
Another difference between these four models is that StrongBaseline performs on the grid-based image features from ResNet-101~\cite{tech:Res101}, while others employ the object-level image features from Faster R-CNN~\cite{tech:RCNN}.

		%
		%
		%
		%
		%
		%
		%
		%

\subsection{Explanation Methods}
We mainly study the attention model explainability from three groups of explanation methods -- \emph{generic attention-based}, \emph{transformer-based}, and \emph{gradient-based} attribution approaches. In particular, generic attention-based methods are suitable for all kinds of attention models, while transformer-based ones are exclusive for transformer architectures. We elaborate the utilized methods in Table \ref{exp_methods}.

Starting with generic attention-based explanation methods, we employ Inherent Attention Explanation (RawAtt)~\cite{attfinding:AttIsNotExp}, Attention $\odot$ Gradient (AttGrad)~\cite{attfinding:2019IsAttIntepret,faithful:ImrpoveFaithfulForTC,transformer:IGAttGrad}, and Attention $\odot$ InputNorm (AttIN)~\cite{attfinding:AttNorm} for all models. Note that in transformer-based architectures, these methods are implemented based on the last attention layer's output following~\cite{transformer:AttGrad_Chefer}.

Regarding the exclusive architecture of transformers (\eg, multi-head attention), we particularly adopt four explanation methods: {Partial Layer-wise Relevance Propagation (PLRP)}~\cite{transformer:LRP_Partial_2019}, {Attention Rollout}~\cite{transformer:AttFlow}, {Transformer Attention Attribution (TransAtt)}~\cite{transformer:LRP_Chefer}, and {Generic Attention Attribution (GenAtt)}~\cite{transformer:AttGrad_Chefer}.

Additionally, in our experiments, we also select two representatives -- Input $\odot$ Gradient (InputGrad)~\cite{exp:DeepLIFT,exp:inputXgrad} and {Integrated Gradients (IG)}~\cite{exp:IG} -- from gradient-based attribution methods for a further comparison.


\begin{table}[t]
	\caption{The faithfulness evaluation metrics utilized in this paper. Appendix \ref{appendix:metrics} provides more details for these metrics.}\smallskip
	\centering
	\resizebox{1\columnwidth}{!}{
		\smallskip\begin{tabular}{l c l}
			\toprule
			Metric & Denoted & Measurement \\
			\midrule
			\multirow{1}{*}{\makecell[l]{AUC-TP}} & \multirow{1}{*}{A $\downarrow$} &  \multirow{1}{*}{Performance perturbation} \\
			
			Sufficiency & S $\downarrow$ &  Confidence perturbation \\
			
			Comprehensiveness & C $\uparrow$ &  Confidence perturbation \\
			
			Rank Correlation  & R $\uparrow$ &  Overall confidence variance \\
			
			Ours Violation Test & V $\downarrow$ &  Confidence perturbation polarity\\
			\bottomrule
		\end{tabular}
	}
	\label{metrics}
\end{table}

\subsection{Faithfulness Evaluation Metrics}
We compare our violation test with existing faithfulness evaluation metrics, while they mainly focus on the importance correlation of explanation methods. Concretely, we experiment with four widely adopted metrics as following.

\textbf{AUC on Threshold Performance (AUC-TP)}~\cite{attfinding:DiagnosticAllProperties}. This metric calculates the AUC score based on the performance perturbation \emph{w.r.t} the feature replacement of top-0, 5, 10, 20, 30, ..., 90\% tokens in the input. 

\textbf{Sufficiency}~\cite{replace:benchmark} measures whether explanations can identify important features, which are adequate for a model to remain confidence on its original predictions. It is calculated as the average of $\Delta {\rm C}$ by sequentially retaining the features in top-5, 10, 20, 50\% sparsity levels.

\textbf{Comprehensiveness}~\cite{replace:benchmark} evaluates if the features assigned with lower weights are unnecessary for the predictions. In analogous to Sufficiency, we compute it as the average of $\Delta {\rm C}$ via sequentially removing the features in the top-5, 10, 20, 50\% sparsity levels.

\textbf{Rank Correlation (Absolute)}~\cite{attfinding:AttIsNotExp,exp:GradCAM} directly calculates the perturbation-based prediction variance for each token, and measures its relationship with the magnitude of explanation weights.

%

\subsection{Replacement Functions}
In the light of out-of-distribution (OOD) problems in perturbation examination~\cite{saliency:RemOveAndRetrain_BeemKim2019}, we utilize three replacement functions as suggested by~\citet{replace:OOD}. 
We provide a brief overview of the involved functions in the following. Note that in this work, all replacements are performed on the token level, \ie, text token or image region.

\textbf{Slice Out}~\cite{replace:benchmark} directly removes specified tokens from the original input, thereby resulting in a shorter length of the new input sequence.

\textbf{Attention Mask}~\cite{attfinding:2019IsAttIntepret} sets the attention weights for removed tokens to 0, such that the removed parts will not be forwarded in the network anymore. 

\textbf{MASK Token}~\cite{replace:Erasure_Jiwei_2016,exp:IG} simply replaces the words with the MASK token. If the token is an image region, we would set it to the zero vector.

In the following sections, all the results are averaged by using the above three replacement functions.

\section{Experimental Results}
Faithfulness violation indicates that an explanation does not truly reflect the impact of input components. In this work, we mainly study the faithfulness violation issue in terms of the polarity consistency, \ie, whether an explanation can correctly identify the impact polarity of features. In particular, we consider the context of attention models due to their wide applications. As described in the previous section, we explore 9 explanation methods across 8 popular model architectures over 6 tasks. 
To sum up, our exploration is to answer the following research questions:
\begin{itemize}
	\item \textbf{RQ1}: Why we need faithfulness violation test? Can existing evaluation metrics reflect the violation issue?
	
	\item \textbf{RQ2}: How existing methods perform on faithfulness? Can any method pass the faithfulness violation test? 
	
	\item \textbf{RQ3}: What factors affect the violation issue?
\end{itemize}

Regarding the first question, we begin by comparing our violation test with existing faithfulness evaluation metrics. We show that all of these metrics are incapable of detecting faithfulness violations. In addition, we further verify the function of our violation test, where the results demonstrate the capability of the test method in detecting violation behavior.  In line with such experiments, for the second question, we conduct a sanity faithfulness evaluation for existing explanation methods. Unexpectedly, the results suggest that almost all approaches are hindered by the violation issue. In particular, the single-polarity methods (\eg, RawAtt and AttIN) perform worst, while the gradient-based explanations (\eg, GenAtt) could greatly alleviate the violation issue. Finally, for the third question,  we study the factors leading to the above issues. Through the experiments, we find that the complexity of model architectures and the ability to identify polarity can be the dominant factors.



\begin{figure}
	[t]
	\centering 
	\includegraphics[width=1\columnwidth]{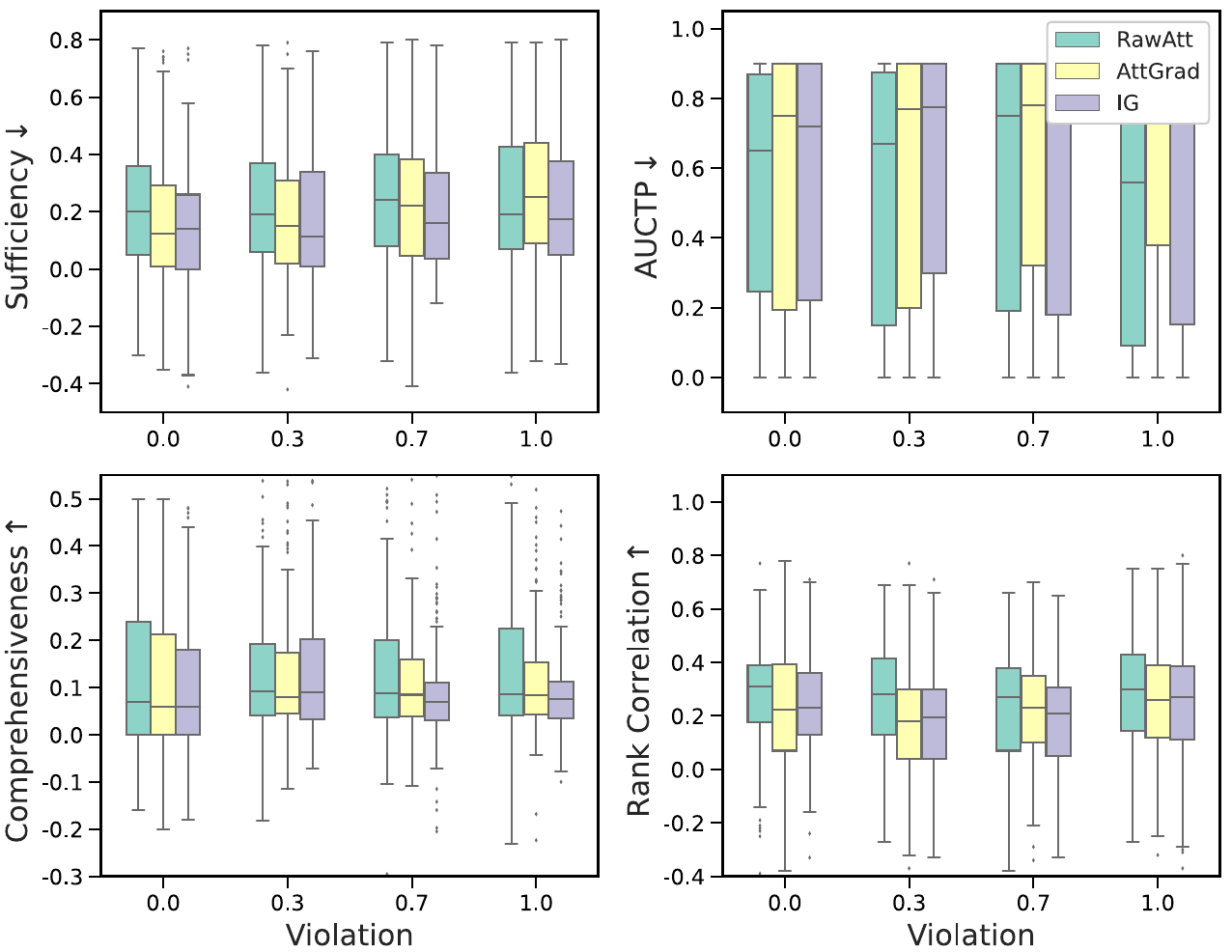} %
	\caption{Faithfulness violation levels ($x$-axis) in correlation to the existing faithfulness evaluation metrics ($y$-axis). The violation level is obtained by the average of our violation results, which denotes worse faithfulness as it increases. The presented explanation methods (RawAtt, AttGrad, and IG) are based on the LXMERT model over the GQA dataset. }
	\label{vio_metric_cpr}
\end{figure}


\begin{figure}
	[t]
	\centering 
	\includegraphics[width=1\columnwidth]{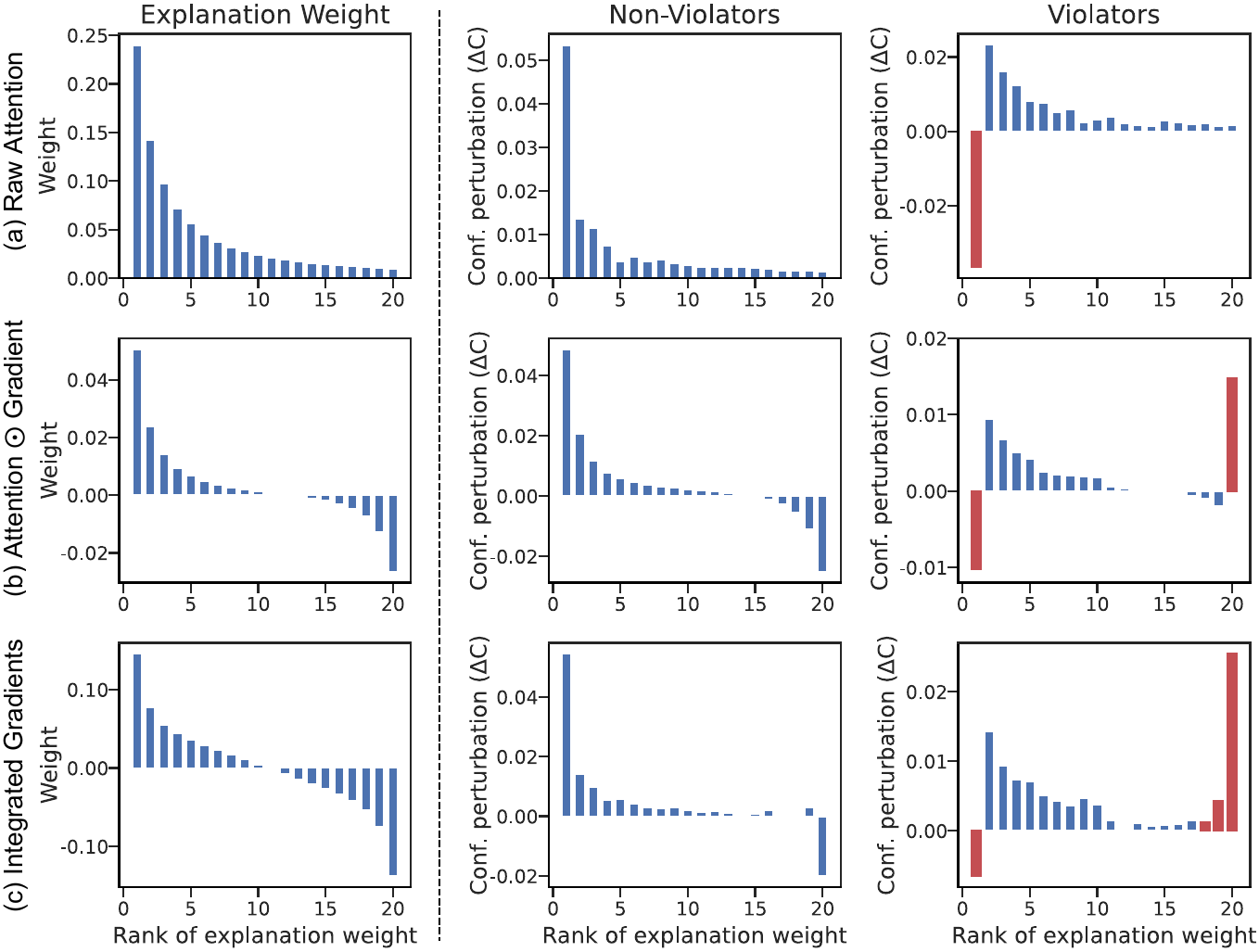} %
	\caption{Comparison between the identified faithfulness non-violators (middle figures) and violators (right ones). The leftmost figures exhibit ordered explanation weights at each rank. The rest figures represent the correlation between explanation weights and perturbation-based input impact ($\Delta {\rm C}$). \emph{Red} bars denote the faithfulness violation in polarity consistency, \ie, the polarity of explanation weight is opposite with the $\Delta {\rm C}$.
	The results are presented on the LXMERT model over the GQA dataset.}
	\label{vio_behavior}
\end{figure}


\subsection{Why violation test is necessary? (RQ1)}
\label{sec_q1:why_vio}



\bfstart{Comparison with existing evaluation metrics} 
We first compare the existing evaluation metrics with our proposed violation test. 
Concretely, we randomly select the same number of data samples at different violation levels, and then calculate their corresponding performance distribution for each evaluation metric (see Figure \ref{vio_metric_cpr}). 
To our surprise, it can be found that at different violation levels, the performance of explanation methods on the same evaluation metric is highly close. 
The correlation plot between the Sufficiency and Violation for the RawAtt can be a concrete example.
As the faithfulness violation level increases, the scores on the Sufficiency, however, do not exhibit an increasing trend, \ie, higher Sufficiency means worse faithfulness. 
Even in the large divergence between 0.0 and 1.0 violation levels, their distribution for the Sufficiency scores is still highly close. In this sense, we can see that existing metrics do not equip capabilities of examining the violation issue, which highlights the necessity of our violation test.

\bfstart{Analysis of violation behavior} To further verify that our test is responsible for the faithfulness violation detection, we compare the behavior of identified faithfulness violators and non-violators (See Figure \ref{vio_behavior}). 
From the results, we can observe that in non-violators, the explanation weights and the measured impact for each input tokens are highly consistent. For example, for raw attention maps, the input tokens with higher positive weights actually show large contribution effects to the model predictions. This behavior is in stark with the detected violators, where the input with the largest weights (in absolute values) tends to present the opposite effects as the polarity indications, \eg, heavy contribution effects are assigned with the most negative weights in Attention $\odot$ Gradient. Such divergence in their behaviors demonstrates the effectiveness of our proposed test for the faithfulness violation examination.

In addition, it can be further observed that the relatively lower explanation weights often exhibit the consistent polarity with the input impact, \eg, inputs with rank-2 positive weights mostly impose contribution effects. This may be the reason why existing explanation metrics perform poorly in violation detection -- their calculation is based on the test of a large proportion of input tokens, such that the polarity violation in the forefront weight can be easily concealed.


\begin{figure}
	[t]
	\centering 
	\includegraphics[width=0.9\columnwidth]{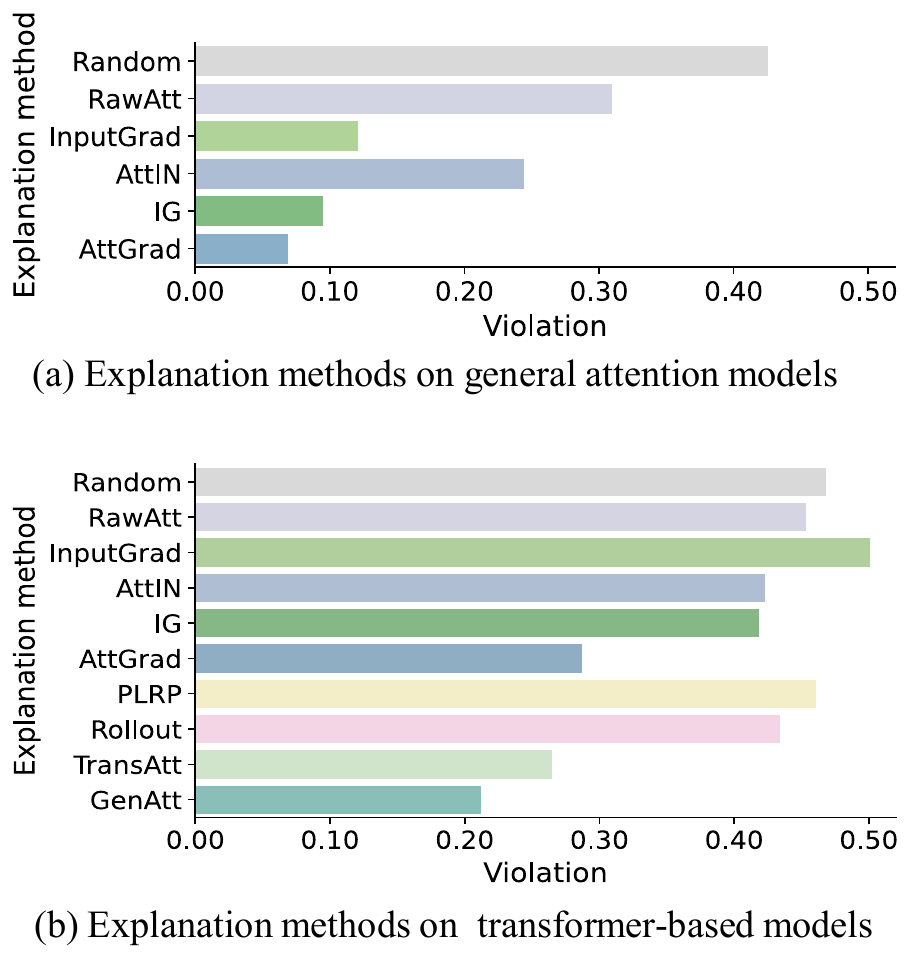} %
	\caption{Faithfulness violation test for the analyzed explanation methods. The results are averaged on adopted models and datasets.}
	\label{eval_vio}
\end{figure}

\begin{figure*}
	[t]
	\centering 
	\includegraphics[width=0.93\textwidth]{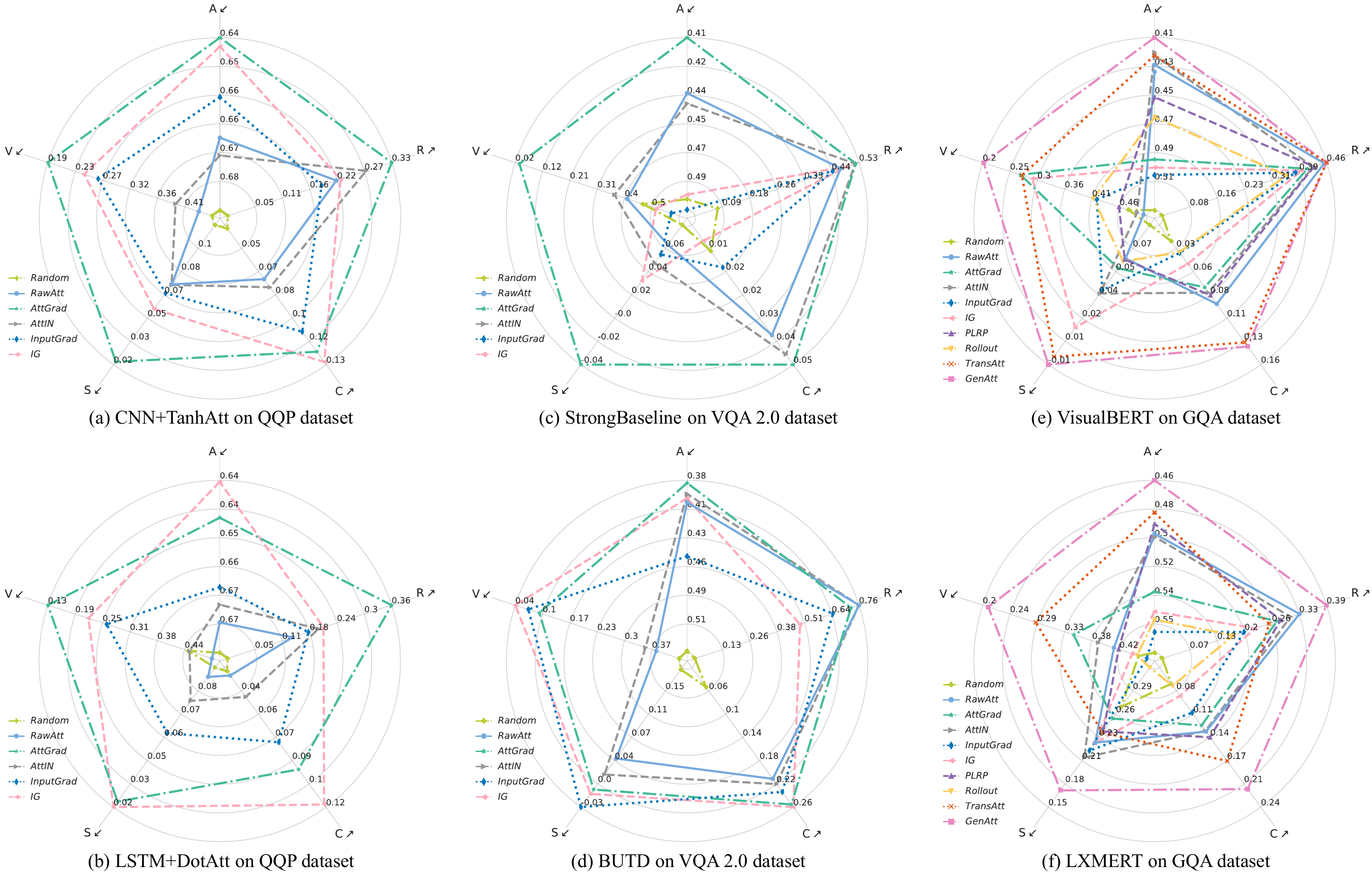} %
	\caption{Faithfulness diagnosis for explanation methods over different architectures, on the QQP, VQA 2.0, GQA datasets. The $\swarrow$ signs indicate the lower value of the corresponding metric is better, while $\nearrow$ denotes the higher value is better. {Table \ref{metrics} exhibits the denotations of metrics.}
	Appendix \ref{appendix:sanity_eval} provides the plots for all models and datasets, and the average evaluation scores for all explanation methods. }
	\label{eval_radar}
\end{figure*}

\subsection{Evaluation results for explanation methods (RQ2)} 
\label{sec_q2:how_exp}
For a complete faithfulness evaluation and comparison, we further examine explanation methods with all faithfulness evaluation metrics on all utilized models and datasets. Due to space constraints, we only provide visualized results for 3 datasets (QQP, VQA 2.0, GQA) in Figure \ref{eval_radar}. 
In particular, the average of violation test measures for explanation methods is illustrated in Figure \ref{eval_vio}.
We provide the main observations in the following:

Firstly, we can observe that all tested explanation methods unexpectedly suffer from the faithfulness violation issue, \ie, violation ratio $>$ 0. While in some cases many methods hardly exhibit the violation behavior, they can still perform worse on other architectures or datasets (See Figure \ref{eval_radar}). For example, the explanation method IG on the BUTD model only shows 4\% violation ratio; whereas the 23\% violation ratio can be drawn from the CNN+TanhAtt model.

Secondly, among the tested attention-based explanation methods, RawAtt and AttIN perform the worst on the violation test. We consider this is due to the fact that their explanation weights are only with one polarity, \ie, positive (See Figure \ref{vio_behavior} for an example of RawAtt weights). 
In this situation, their single-polarity weights do not equip the capability to identify whether the feature impact is positive or negative, thereby leading to the high faithfulness violation (We provide more discussion in Section \ref{section:factors}). Moreover, we can further observe that even these two methods are heavily hindered by the violation issue, they commonly do not show poor performance on the other evaluation metrics (See Figure \ref{eval_radar}). For instance, in BUTD and LXMERT models, while RawAtt and AttIN present severe violations (about 40\%), they can still show promising performance on the AUCTP (A) and Sufficiency (S), which is even comparable or superior to the AttGrad (Note that AttGrad performs much better than them on the violation test). This finding helps us further deduce that the execution of violation test is imperative for a sanity faithfulness evaluation.

Lastly, through the results, the form of the Attention $\odot$ Gradient performs the best almost on all architectures and datasets. In detail, for the general attention models (\eg, CNN+TanhAtt or StrongBaseline), the AttGrad method often shows the best performance on all metrics. Likewise, for the transformer-based models, the another explanation approach GenAtt behaves better. 


\subsection{Factors affecting faithfulness violation issue (RQ3)}
\label{section:factors}
The previous section shows that the existing explanation methods mostly violate the faithfulness property in terms of polarity consistency across models and datasets, while the gradient-based explanations can alleviate it to some extent. 
Herein, we take a step further to study the potential factors leading to the faithfulness violation issue. 

\begin{figure}
	[t]
	\centering 
	\includegraphics[width=0.99\columnwidth]{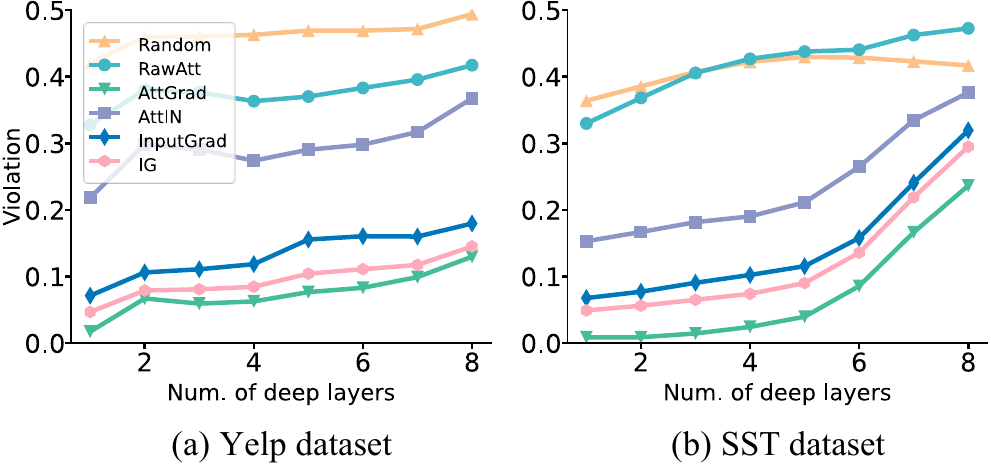} %
	\caption{Faithfulness violation in relation to the number of non-linear deep layers. The results for Yelp and SST datasets are averaged on the four utilized attention models: CNN+TanhAtt, LSTM+TanhAtt, CNN+DotAtt, and LSTM+DotAtt.}
	\label{factors_layer}
\end{figure}

\textbf{Do deeper layers hinder the explanation faithfulness?} As shown before in Figure \ref{eval_vio} and Figure \ref{eval_radar}, the explanation methods which perform well on the general attention models (\eg, CNN+DotAtt) often exhibit a worse faithfulness on the transformer-based models, \eg, an increase of 14\% violation ratio in AttGrad. In the light of the prior studies~\cite{exp:DeepLIFT,exp:IG,saliency:WhyBPFail_ICML2020}, we hypothesize that the divergence in number of stacked deep layers \emph{w.r.t} these two architectures may be the reason. To this end, we study the faithfulness of explanation methods when the number of deep layers increases on the Yelp and SST datasets (See Figure \ref{factors_layer}). 
Through the results, a positive correlation between the layer numbers and the faithfulness violation can be observed on almost all explanation methods over both datasets, which confirms our hypothesis in the negative effects of deep layers. Besides, as model architecture grows in complexity and learning functions, this experiment further poses an imperative demand in developing robust explanation methods.

\textbf{Is it crucial to equip capability of identifying impact polarity?} In our previous findings, we show that AttIN and RawAtt are weak in identifying impact polarity because of their invariably positive weights. To verify the reason, we compare the explanation faithfulness of Attention $\odot$ Gradient with the raw attention. 
As illustrated in Table \ref{factors_att}, we ablate the formulation of Attention $\odot$ Gradient (${\alpha} \odot \nabla\alpha$) into two forms: 1) remaining only the sign of gradients (${\alpha}~\odot~{\rm sign}(\nabla\alpha)$), and 2) remaining merely the absolute value of gradients (${\alpha} \odot |\nabla\alpha|$). 
By comparison with the raw attention explanations, we can observe that even with the signs of gradients only, the faithfulness still shows a large improvement over the raw attention. In addition, compared to the intact gradient information, the violation issue would also become more severe if we merely incorporate the absolute value of gradients. These observations spotlight the importance of capability to identify the impact polarity. 

\begin{figure}
	[t]
	\centering 
	\includegraphics[width=0.93\columnwidth]{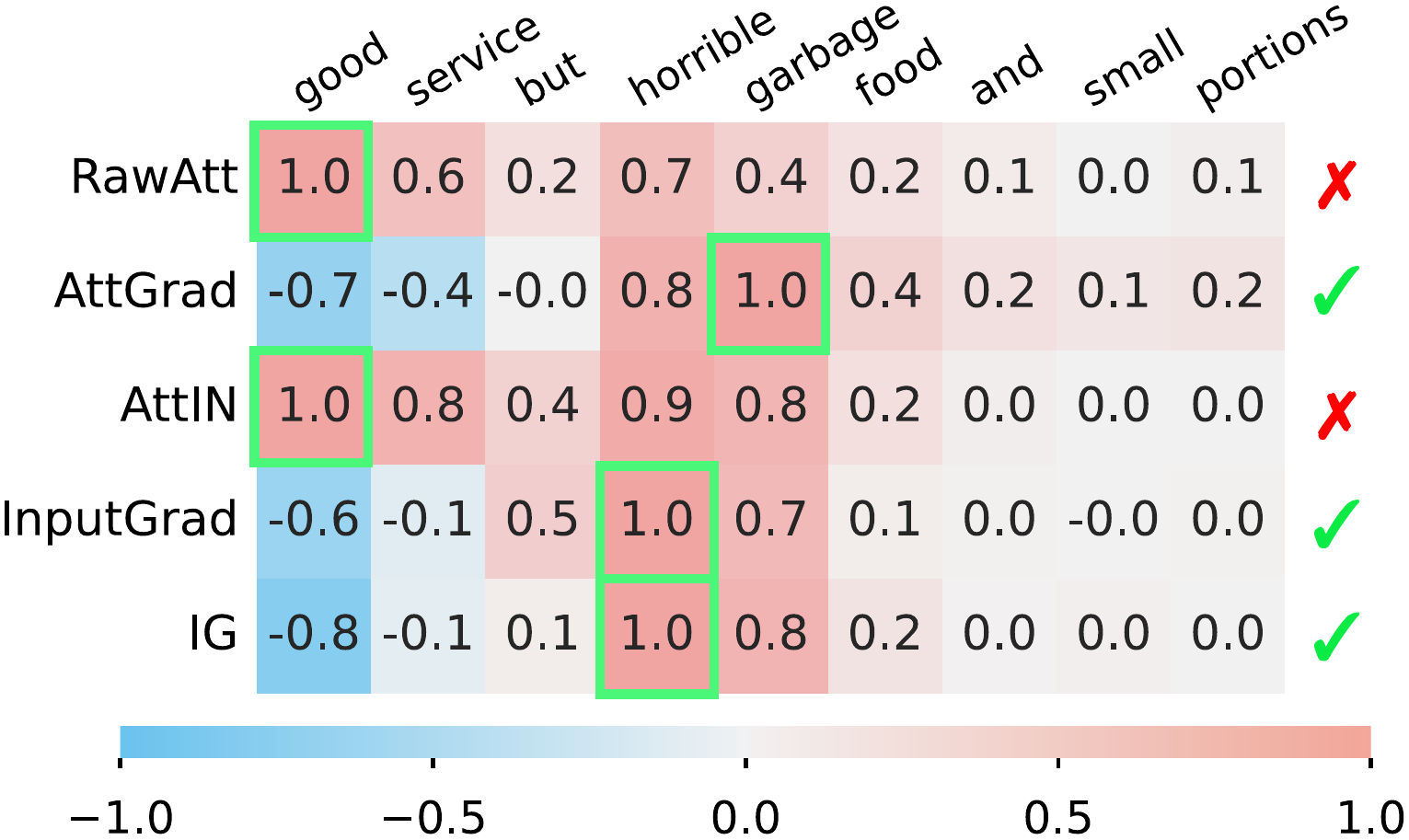} %
	\caption{{Normalized} explanations for an example from the sentiment analysis task over the Yelp dataset. \emph{Negative} sentiment is the model prediction for this example. Red boxes represent that indications of explanation weights are contribution effects to the model prediction. Blue boxes denote the indications are suppression effects. \textcolor{darkpastelgreen}{\cmark}: Non-violation. \textcolor{red}{\xmark}: Violation. }
	\label{examples}
\end{figure}

Furthermore, we also provide an example in Figure \ref{examples} to illustrate the above point. 
As can be seen, the given example consists of two parts with opposite sentiments (\ie, positive from ``good service'' and negative from ``horrible garbage''), which impose the contrary impacts on the model prediction. Suffered from the weakness of single-polarity weights, however, AttIN and RawAtt only can assign the same polarity for these two parts of words, which is therefore highly misleading and heavily violate the faithfulness.





\begin{table}[t]
	\caption{Faithfulness violation ratios for ablations of attention-based explanation methods. ${\alpha}$ denotes the raw attention weights.}\smallskip
	\centering
	\resizebox{0.8\columnwidth}{!}{
		\smallskip\begin{tabular}{c c c c}
			\toprule
			Method & Yelp & AgNews & VQA 2.0 \\
			\midrule
			${\alpha}$ & 0.31 & 0.28 & 0.40 \\
			\cmidrule(lr){2-4}
			${\alpha} \odot \nabla\alpha$ & \textbf{0.02} & \textbf{0.03} & \textbf{0.06} \\
			${\alpha} \odot |\nabla\alpha|$ & 0.15 & 0.07 & 0.25 \\
			${\alpha} \odot {\rm sign}(\nabla\alpha)$ & 0.16 & 0.18 & 0.27\\
			\bottomrule
		\end{tabular}
	}
	\label{factors_att}
\end{table}

\section{Discussions and Conclusions}

In this paper, we have presented the faithfulness violation view for efficient comparison of attention-based explanation methods. We have demonstrated that the polarity consistency defined with concrete mathematical formulation provides the reasonable evaluation criterion on the explainability of attention models. 
To facilitate the understanding of the faithfulness violation test, we have shown that our method is essential for the faithfulness evaluation in comparison with existing metrics (Section \ref{sec_q1:why_vio}). 
Despite remarkable progress has been made in attention modeling, our results unexpectedly demonstrate that most tested explanation methods are hindered by the faithfulness violation issue in terms of the polarity consistency (Section \ref{sec_q2:how_exp}). 
Nevertheless, this does not necessarily mean that 
the attention-based explanations examined are not meaningful, though the research has not yet reached the level of maturity. 
To gain useful insights in designing better models and explanations, we have studied the factors affecting the violation issue (Section \ref{section:factors}), which suggests that the complexity in model architectures and the capability of identifying polarity can be the dominant factors. 
These impressive results highlight not only the current state of attention-based explanations, but also the challenges in designing and benchmarking future explanation methods.

Given the diversity of design philosophies, it is inevitably non-trivial to generalize across model architectures and tasks. Even though, the capability of identifying impact polarity should not be ignored in designing an explanation method. Without such capability, the explanations cannot tell us whether an input component would contribute to the model prediction or not, which can easily mislead us in understanding the rationales behind model decisions. While incorporating the gradient-based explanation is not a flawless approach, our findings reveal that it does provide some clues and helps in polarity identification. 

{We acknowledge that the confidence perturbation can be the limitation in our faithfulness violation test, thus depressing the evaluation results to some extent. However, we would like to highlight again that our experiments are conducted by using several replacement functions, which are verified as promising solutions for OOD problem in perturbation. In addition, while there can be model noises (\eg, biases) affecting violation identification, such errors are indeed unavoidable in current explanation evaluations, and our experimental results (\eg, Figure \ref{vio_behavior}) have also shown the potential of our tested method in faithfulness examination.}

It is worth mentioning that the attention-based explanation methods discussed so far are constrained in an ideal narrow scope, facilitating the in-depth and focused discussion. However, there is a wider variety of applications and an enormous demand in attention model explanations. 
How to adapt the faithfulness violation to quantify the quality of explanation methods has great potential in improving model explainability. 
Lastly, we hope that we have provided further motivation to the community to consider the faithfulness violation as the design criteria in explanation methods and explainable attention models.
We also envision that this is the most considerable benefit eventually lies.

{
\section*{Acknowledgements}
This work is supported by the National Natural Science Foundation of China under 62022002, the Hong Kong Research Grants Council General Research Fund (GRF) under Grant 11203220, the Hong Kong Innovation and Technology Commission (InnoHK Project CIMDA), the CityU Teaching Start-up Grant 6000801, the CityU Strategic Interdisciplinary Research Grant (SIRG) 7020055, and the CityU New Research Initiatives/Infrastructure Support from Central (APRC 9610528). We thank the ICML reviewers for their valuable comments on improving this presentation.
}

\bibliography{mybib}
\bibliographystyle{icml2022}

\newpage
\appendix
\onecolumn
\section{Tasks and datasets}
\label{appendix:datasets}
\begin{itemize}
	\item \textbf{Sentiment Analysis}. We use Stanford Sentiment Treebank (SST)~\cite{dataset:SST} and Yelp\footnote{www.yelp.com/dataset\_challenge} datasets. In this task, the trained models are required to predict sentiment for each sentence from five levels (1 for most negative and 5 for most positive). 
	
	\item \textbf{Topic Classification}. We utilize  AGNews\footnote{http://www.di.unipi.it/~gulli/AG\_corpus\_of\_news\_articles.html} and 20News\footnote{http://qwone.com/~jason/20Newsgroups} datasets, where the target is to predict categories (\eg, business) for each article. In particular, AGNews contains 496,835 articles from 2000+ sources, while 20News consists of ~20,000 samples.
	
	\item \textbf{Paraphrase Detection}. We adopt the Quora Question Paraphrase (QQP) dataset~\cite{dataset:QQP} following ~\cite{attfinding:LSTM_Concinity}. In this task, models are trained  to determine the semantic equivalence of a pair of questions.
	
	\item \textbf{Natural Language Inference}. We utilize the SNLI~\cite{dataset:SNLI}, which includes about 570K human-written English sentence pairs. In this task, models are trained for identifying textual entailment within sentence pairs.
	
	\item \textbf{Question Answering}. We make use of the bAbI-1 dataset~\cite{dataset:bAbI} here. In this dataset, correctly answering each question requires a supporting fact from the context.
	
	\item \textbf{Visual Question Answering (VQA)}. We use VQA 2.0~\cite{dataset:VQA_v2} and GQA~\cite{dataset:GQA} datasets, which include about 410K and 950K training samples, respectively. In this task, models are required to understand both images and texts for correctly answering questions.
\end{itemize}



\section{Details of explanation methods}
\label{appendix:expmethod}
We mainly study the attention model explainability from three groups of explanation methods -- \emph{generic attention-based}, \emph{transformer-based}, and \emph{gradient-based} attribution approaches. In detail, the following methods are utilized.

\subsection{Generic attention-based explanation methods}
Starting from generic attention-based explanation methods, we employ them in all attention models.

\textbf{Inherent Attention Explanation (RawAtt)}~\cite{attfinding:AttIsNotExp} is the direct output from attention mechanisms. In \emph{AddAtt}, for example, it is calculated by ${\alpha} = {\rm softmax}(\mathbf{w}_1^T {\rm tanh}(\mathbf{W}_2{\rm K} + \mathbf{W}_3{\rm Q}))$, where ${\rm K}$ and ${\rm Q}$ are query and keys. $\mathbf{w}_1$, $\mathbf{W}_2$, and $\mathbf{W}_3$ are learned parameters.

\textbf{Attention $\odot$ Gradient (AttGrad)}~\cite{attfinding:2019IsAttIntepret,faithful:ImrpoveFaithfulForTC,transformer:IGAttGrad} scales attention weights with the corresponding gradients, denoted as
${\alpha} \odot \nabla\alpha$, where $ \nabla\alpha = \frac{\partial \hat{y}}{\partial \alpha}$.

\textbf{Attention $\odot$ InputNorm (AttIN)}~\cite{attfinding:AttNorm} incorporates another factors, input vectors, in the attention-based explanations. \emph{InputNorm} corresponds to the norm of transformed input vectors, $||v(x)||$, in attention modules, where $v(\cdot)$ can be the value mappings in transformers.

\subsection{Transformer-based explanation methods}
Regarding the exclusive architecture of transformers (\eg, multi-head attention), we further adopt explanation methods for transformer-based models in the following.

\textbf{Partial Layer-wise Relevance Propagation (PLRP)}~\cite{transformer:LRP_Partial_2019} computes the LRP-based scores~\cite{exp:LRP} for each attention head in the last layer\footnote{LRP is a method to compute contribution  scores for neurons. }, and then average them across heads as explanations. 

\textbf{Rollout}~\cite{transformer:AttFlow} builds upon the assumption that attention weights can be combined linearly following paths along the pairwise attention graph.

\textbf{Transformer Attention Attribution (TransAtt)}~\cite{transformer:LRP_Chefer} is another form of LRP-based explanation method. It calculates LRP scores for each attention heads, and then integrates such scores with attention gradients throughout the attention graph. We simplify it as $\nabla\alpha \odot {\rm R}^{\alpha}$, where ${\rm R}^{\alpha}$ is the LRP scores for the corresponding head.

\textbf{Generic Attention Attribution (GenAtt)}~\cite{transformer:AttGrad_Chefer} provides general solution to explain different transformer architectures. GenAtt combines attention weights with corresponding gradients for all layers based on the devised adaptation strategies, and therefore can be deemed as another form of the ${\alpha} \odot \nabla\alpha$.

\subsection{Gradient-based attribution methods}
Besides the above methods, we also select two representatives from gradient-based approaches for a comparison:

\textbf{Input $\odot$ Gradient (InputGrad)}~\cite{exp:DeepLIFT,exp:inputXgrad} calculates the element-wise
product between the input $x$ and the corresponding gradient: ${x} \odot \frac{\partial \hat{y}}{\partial x}$. 

\textbf{Integrated Gradients (IG)}~\cite{exp:IG} alleviates gradient saturation issue by aggregating input gradients along the path between a baseline input ($x'$) to the original input ($x$), defined as
$(x-x') \times \int_{0}^{1} \frac{\partial f(x' + \gamma(x-x'))}{\partial x} d\gamma $.

{

	
}

{\section{Details of faithfulness evaluation metrics}
\label{appendix:metrics}

\textbf{AUC-TP $\downarrow$}~\cite{attfinding:DiagnosticAllProperties}. The calculation of this metric is based on the AUC score of performance perturbation. Concretely, we first sequentially mask the top-0, 5, 10, 20, ..., 90\% input tokens \emph{w.r.t.} explanation weights, to obtain a set of model performance $S = \{s^0, s^{5}, s^{10}, s^{20}, ..., s^{90}\}$. Then based on the mask levels, we calculate the area under threshold-performance curve, defined as
${\rm AUC} ([0, 5\%, 10\%, 20\%, ..., 90\%], S)$.

\textbf{Sufficiency $\downarrow$}~\cite{replace:benchmark} measures whether important features identified by the explanation method are adequate to remain confidence on the original predictions. Formally, 
\begin{equation}
{\rm Sufficiency} = \frac{1}{B} \sum_{k \in B}
f(x)_{\hat{y}} - f(x_{:k\%})_{\hat{y}},
\end{equation}
where $B = \{5, 10, 20, 50 \}$, representing sparsity levels of explanation weights in order of decreasing explanation weights.

\textbf{Comprehensiveness $\uparrow$}~\cite{replace:benchmark} evaluates if the features assigned lower weights are unnecessary for the predictions. Formally,
\begin{equation}
{\rm Comprehensiveness} = \frac{1}{B} \sum_{k \in B}
f(x)_{\hat{y}} - f(x \backslash x_{:k\%})_{\hat{y}},
\end{equation}
where $B = \{5, 10, 20, 50 \}$, denoting sparsity levels of explanation weights in order of decreasing explanation weights.

\textbf{Rank Correlation (RC) $\uparrow$}~\cite{attfinding:AttIsNotExp,exp:GradCAM} measures the monotonic correlation between the explanation weights and the feature importance. The calculation of this metric is based on the Spearman's Rank Correlation Coefficient, \ie, $\rho(\cdot)$. Formally,
\begin{equation}
	{\rm Rank\mbox{-}Correlation} = \rho (\hat{e}, p) ,
\end{equation}
where $\hat{e}$ denotes the absolute explanation weights in the descending order, and $p$ represents the corresponding measured feature importance, \ie, 
$p = \big[\sum_{i}|f(x)_{y_i} - f(x \backslash x_{:1})_{y_i}|,
\sum_{i}|f(x)_{y_i} - f(x \backslash x_{:2})_{y_i}|,
 ..., 
 \sum_{i}|f(x)_{y_i} - f(x \backslash x_{:N})_{y_i}| 
 \big]$.
}


\section{Sanity faithfulness evaluation for explanation methods}
\label{appendix:sanity_eval}

\begin{table}[bh]
	\caption{Average faithfulness evaluation scores for explanation methods on all utilized datasets and model architectures.}\smallskip
	\centering
	\resizebox{0.6\columnwidth}{!}{
		\smallskip\begin{tabular}{l c c c c c}
			\toprule
			\multirow{2}{*}{Method} & \multicolumn{5}{c}{Faithfulness Properties} \\
			\cmidrule(){2-6}
			& AUCTP$\downarrow$ & Violation$\downarrow$ & Suf.$\downarrow$ & Comp.$\uparrow$ & RC$\uparrow$\\
			\midrule 
			\multicolumn{6}{c}{\emph{Explanation for General Attention Models}} \\
			
			Random	&0.570	&0.425	&0.226	&0.060	&-0.030\\
			RawAtt	&0.422	&0.309	&0.105	&0.248	&0.411\\
			InputGrad	&0.386	&0.121	&0.033	&0.310	&0.437\\
			AttIN	&0.402	&0.244	&0.089	&0.257	&0.478\\
			IG	& \textbf{0.345}	&0.095	&\textbf{-0.004}	&\textbf{0.370}	&0.470\\
			AttGrad	&0.352	&\textbf{0.069}	&0.018	&0.359	&\textbf{0.535}\\
			\midrule
			
			\multicolumn{6}{c}{\emph{Explanation for Transformer-based models}} \\
			Random	&0.581	&0.468	&0.189	&0.028	&0.000\\
			RawAtt	&0.477	&0.454	&0.150	&0.136	&0.360\\
			InputGrad	&0.560	&0.501	&0.159	&0.090	&0.321\\
			AttIN	&0.472	&0.423	&0.146	&0.146	&0.136\\
			IG	&0.551	&0.419	&0.139	&0.120	&0.318\\
			AttGrad	&0.538	&0.287	&0.145	&0.127	&0.131\\
			PLRP	&0.471	&0.461	&0.153	&0.148	&0.301\\
			Rollout	&0.533	&0.434	&0.190	&0.073	&0.105\\
			TransAtt	&0.458	&0.265	&0.123	&0.194	&0.299\\
			GenAtt	&\textbf{0.451}	&\textbf{0.212}	&\textbf{0.093}	&\textbf{0.318}	&\textbf{0.434}\\
			
			\bottomrule
		\end{tabular}
	}
	\label{tab:trans_faith_properties}
\end{table}

\begin{figure*}
	[bh]
	\centering 
	\includegraphics[width=0.99\textwidth]{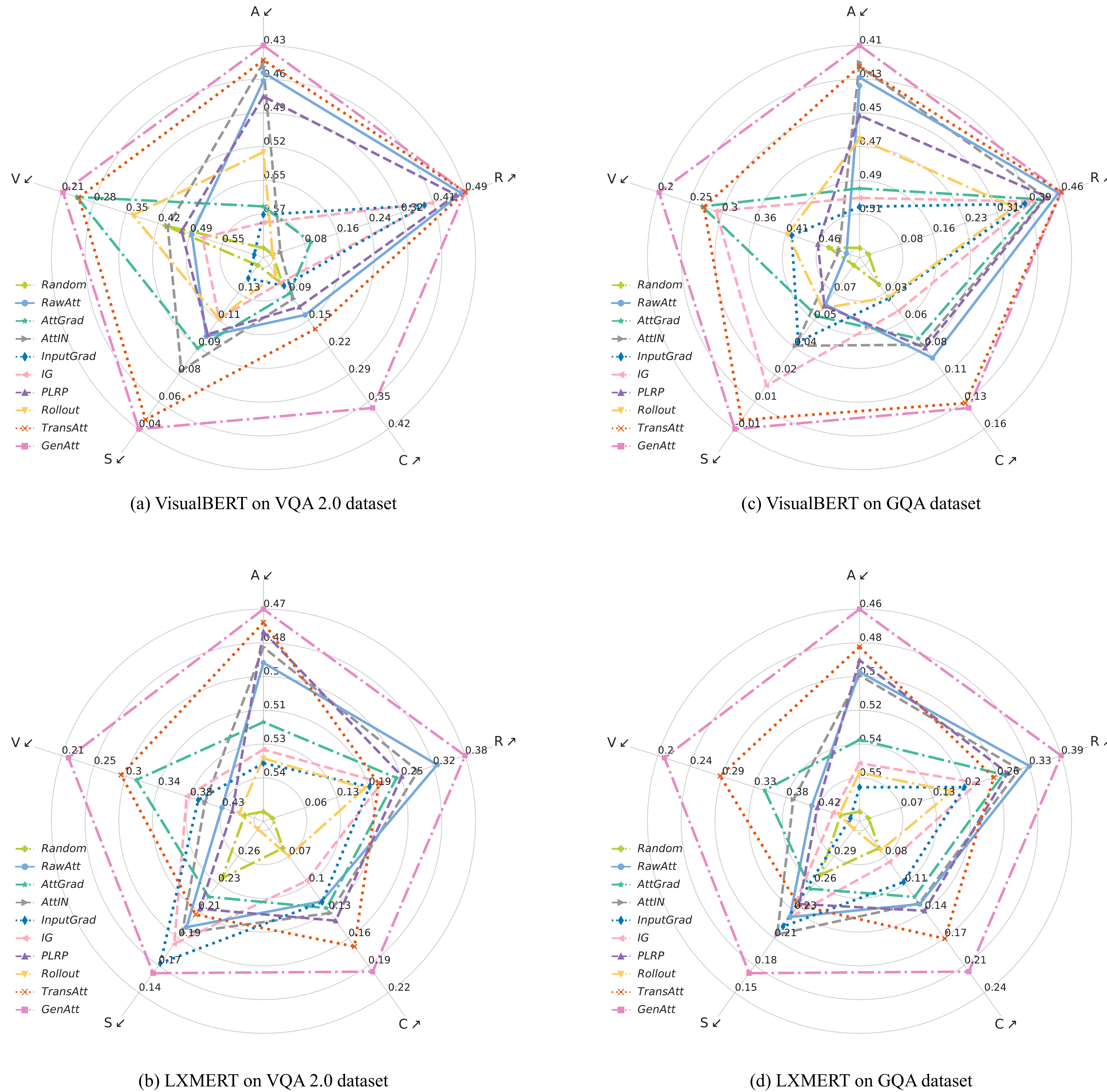} %
	\caption{Faithfulness diagnosis for all explanation methods over different attention-based architectures on VQA 2.0 and GQA datasets.}
	\label{radar_appendix0}
\end{figure*}

\begin{figure*}
	[t]
	\centering 
	\includegraphics[width=0.95\textwidth]{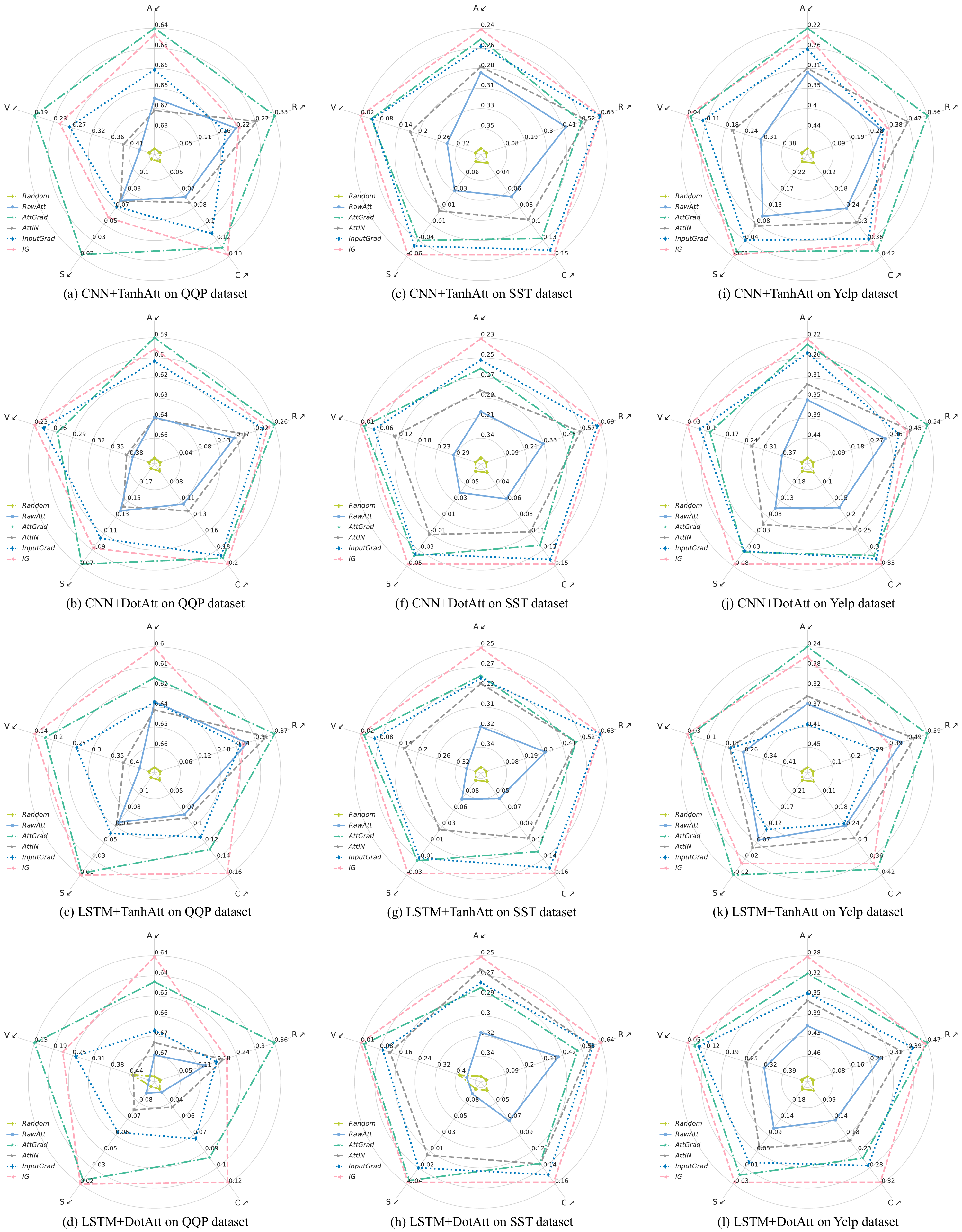} %
	\caption{Faithfulness diagnosis for all explanation methods over different attention-based architectures on QQP, SST, and Yelp datasets.}
	\label{radar_appendix1}
\end{figure*}
\newpage

\begin{figure*}
	[t]
	\centering 
	\includegraphics[width=0.95\textwidth]{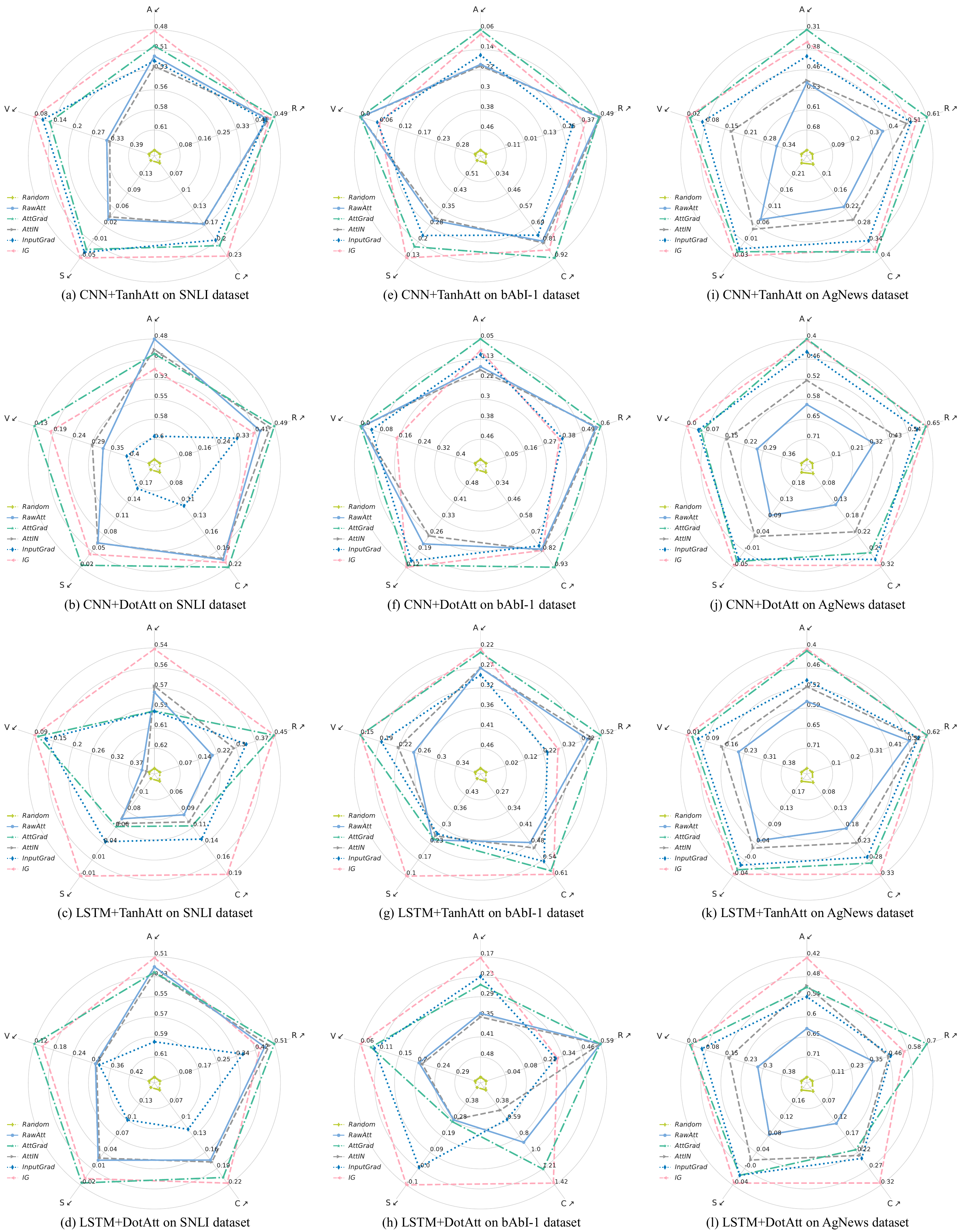} %
	\caption{Faithfulness diagnosis for all explanation methods over different attention-based architectures on the SNLI, bAbI-1, and AgNews datasets.}
	\label{radar_appendix2}
\end{figure*}
\newpage

\begin{figure*}
	[t]
	\centering 
	\includegraphics[width=1\textwidth]{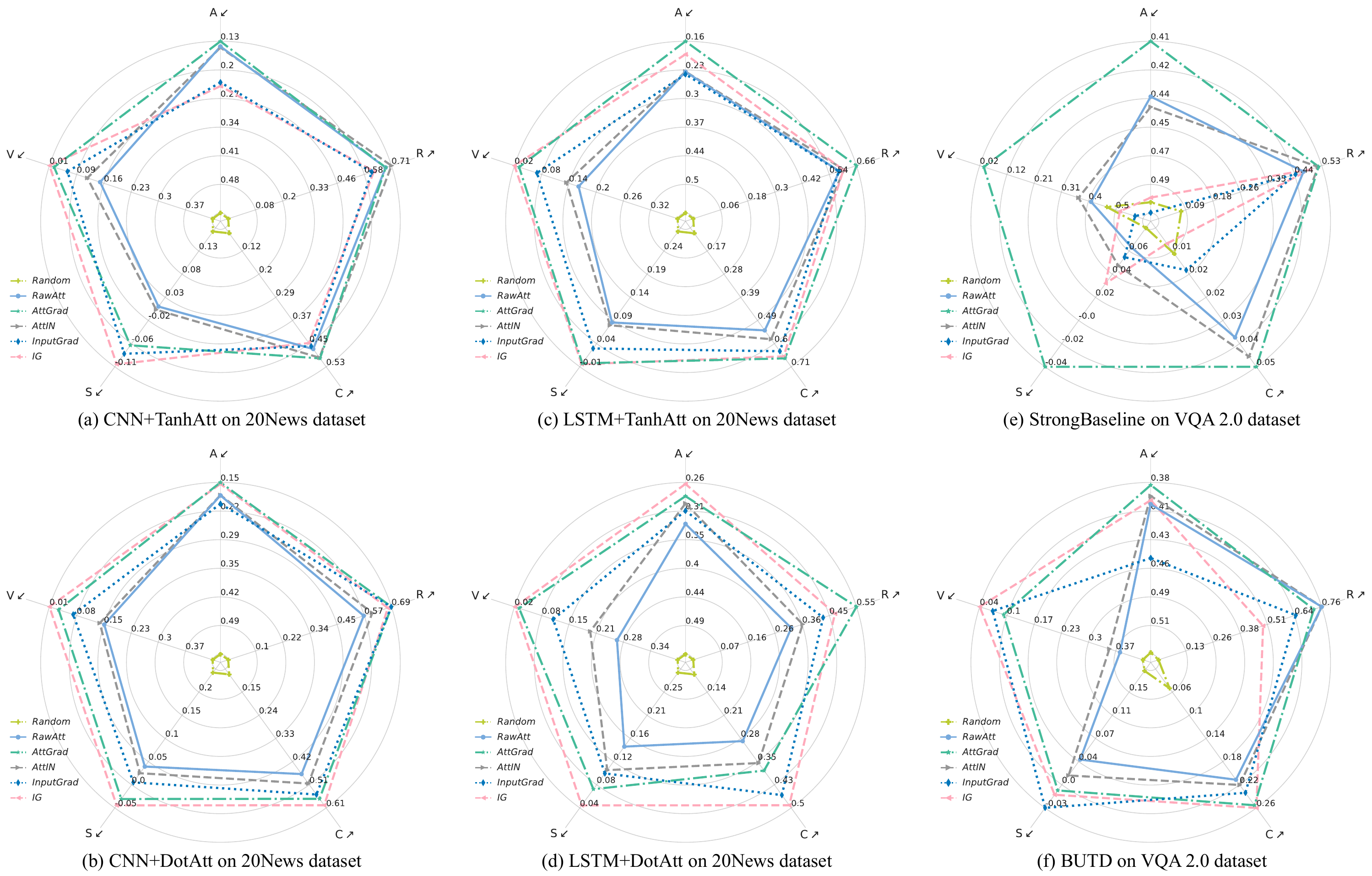} %
	\caption{Faithfulness diagnosis for all explanation methods over different attention-based architectures on 20News and VQA 2.0 datasets.}
	\label{radar_appendix3}
\end{figure*}

\end{document}